\documentclass{ieeetj}
\let\labelindent\relax
\usepackage{cite}
\usepackage{amsmath,amssymb,amsfonts}
\usepackage{algorithmic}
\usepackage{graphicx,color}
\usepackage{textcomp}
\usepackage{xcolor}
\usepackage{enumitem}
\usepackage{hyperref}
\usepackage{url}
\usepackage[labelfont=bf,font=scriptsize]{caption}
\usepackage{gensymb}
\usepackage{multirow}
\usepackage[compact]{titlesec}
    \titlespacing{\section}{0pt}{2ex}{1ex}
    \titlespacing{\subsection}{0pt}{1ex}{0ex}
    \titlespacing{\subsubsection}{0pt}{0.5ex}{0ex}
\usepackage{booktabs} 
\usepackage{tabularx}
\usepackage{comment}
\usepackage{algorithm,algorithmic}
\def\BibTeX{{\rm B\kern-.05em{\sc i\kern-.025em b}\kern-.08em
    T\kern-.1667em\lower.7ex\hbox{E}\kern-.125emX}}
\AtBeginDocument{\definecolor{tmlcncolor}{cmyk}{0.93,0.59,0.15,0.02}\definecolor{NavyBlue}{RGB}{0,86,125}}

\def\OJlogo{\vspace{-4pt}$<$Society logo(s) and publication title will appear here.$>$}
\def\seclogo{\vspace{10pt}$<$Society logo(s) and publication title will appear here.$>$}

\def\authorrefmark#1{\ensuremath{^{\textbf{#1}}}}

\begin{document}

\twocolumn[{%
\vspace{20mm}
{ \large
\begin{itemize}[leftmargin=2.5cm, align=parleft, labelsep=2cm, itemsep=4ex,]

\item[\textbf{Citation}]{K. Kokilepersaud, M. Prabhushankar, Y. Yarici, G. AlRegib, and A. Mostafa, "Exploiting the Distortion-Semantic Interaction in Fisheye Data" in \textit{Open Journal of Signals Processing}, 2023.}

\item[\textbf{Review}]{Date of Publication: May 1st 2023}

\item[\textbf{Codes}]{\url{https://github.com/olivesgatech/SupCon_Ford.git}}

\item[\textbf{Bib}]  {@inproceedings\{kokilepersaud2023exploiting,\\
    title=\{Exploiting the Distortion-Semantic Interaction in Fisheye Data\},\\
    author=\{K. Kokilepersaud, M. Prabhushankar, Y. Yarici, G. AlRegib, and A. Mostafa\},\\
    booktitle=\{Open Journal of Signals Processing\},\\
    year=\{2023\}\}}

%\item[\textbf{Copyright}]{\textcopyright 2022 IEEE. Personal use of this material is permitted. Permission from IEEE must be obtained for all other uses, in any current or future media, including reprinting/republishing this material for advertising or promotional purposes,
%creating new collective works, for resale or redistribution to servers or lists, or reuse of any copyrighted component
%of this work in other works.}

\item[\textbf{Contact}]{
\{kpk6, mohit.p, alregib\}@gatech.edu\\
\url{https://ghassanalregib.info/}\\}
\end{itemize}
}}]
\newpage

\title{Exploiting the Distortion-Semantic Interaction in Fisheye Data}

\author{Kiran Kokilepersaud\authorrefmark{1}, Student Member, IEEE, Mohit Prabhushankar\authorrefmark{1}, Member, IEEE, Yavuz Yarici\authorrefmark{1}, Ghassan AlRegib\authorrefmark{1}, IEEE Fellow, Armin Parchami\authorrefmark{2}}
\affil{OLIVES Lab at the Center for Signals and Information Processing, Georgia Institute of Technology, Atlanta, GA 30332-0250 USA}
\affil{Ford Motor Company, Dearborn, MI 48126 USA}

\corresp{Corresponding author: Kiran Kokilepersaud (email: kpk6@gatech.edu)}
\authornote{This work was supported through the Ford-Georgia Tech Alliance Program.}

\begin{abstract}

In this work, we present a methodology to shape a fisheye-specific representation space that reflects the interaction between distortion and semantic context present in this data modality. Fisheye data has the wider field of view advantage over other types of cameras, but this comes at the expense of high radial distortion. As a result, objects further from the center exhibit deformations that make it difficult for a model to identify their semantic context. While previous work has attempted architectural and training augmentation changes to alleviate this effect, no work has attempted to guide the model towards learning a representation space that reflects this interaction between distortion and semantic context inherent to fisheye data. We introduce an approach to exploit this relationship by first extracting distortion class labels based on an object's distance from the center of the image. We then shape a backbone's representation space with a weighted contrastive loss that constrains objects of the same semantic class and distortion class to be close to each other within a lower dimensional embedding space. This backbone trained with both semantic and distortion information is then fine-tuned within an object detection setting  to empirically evaluate the quality of the learnt representation. We show this method leads to performance improvements by as much as 1.1\% mean average precision over standard object detection strategies and .6\% improvement over other state of the art representation learning approaches. 
\end{abstract}


\maketitle

\section{Introduction}
Autonomous vehicles (AV) have the potential to change existing transportation systems. However, one major concern is the interaction between their acquisition sensors (cameras) and their deep learning based decision algorithms. This concern exists because perception decisions made by an autonomous vehicle is dependent on the quality of the data they receive from the surrounding environment. In particular, camera setups with a wider field of view are attractive due to their ability to capture a more holistic representation of the entire scene. For this reason, fisheye camera lenses are gaining attention as the main vision sensor on these AV systems due to their effective receptive field of 180 degrees \cite{kumar2022surround}. As a result of this advantage, fisheye cameras have seen widespread adoption in common vehicle settings such as parking assistance \cite{hughes2009wide} and automated parking \cite{heimberger2017computer}. Despite their usage in diverse applications, fisheye cameras come with the unique challenge of exhibiting radial distortion as a function of distance from the center of the image. Analysis into the acquisition process of these cameras has shown that this distortion is an inherent consequence of projecting the hemispherical lens geometry onto a 2D plane \cite{miyamoto1964fish}. A naive solution to this problem would be to simply apply a transformation that would rectify the distortion. However, it has been shown \cite{kumar2020unrectdepthnet} that these types of approaches introduce artifacts at the edges and reduce the overall field of view of the image. 

%Previous work \cite{kumar2022surround} has emphasized how the use of four fisheye cameras can form a network with overlap that provides a 360 degree vision of the near field area around a car. 
\begin{comment}
\begin{figure}[t]
\centering
\includegraphics[scale=.35]{FIgures/edge_center_comparison.jpg}

\caption{An example of a standard fisheye image. It is observed that images of the same car class exhibit significantly different distortion characteristics depending on their location within the image space. }

\label{fig: edge_center_analysis}
\end{figure}
\end{comment}

\begin{figure*}[t]
\centering
\includegraphics[scale=.25]{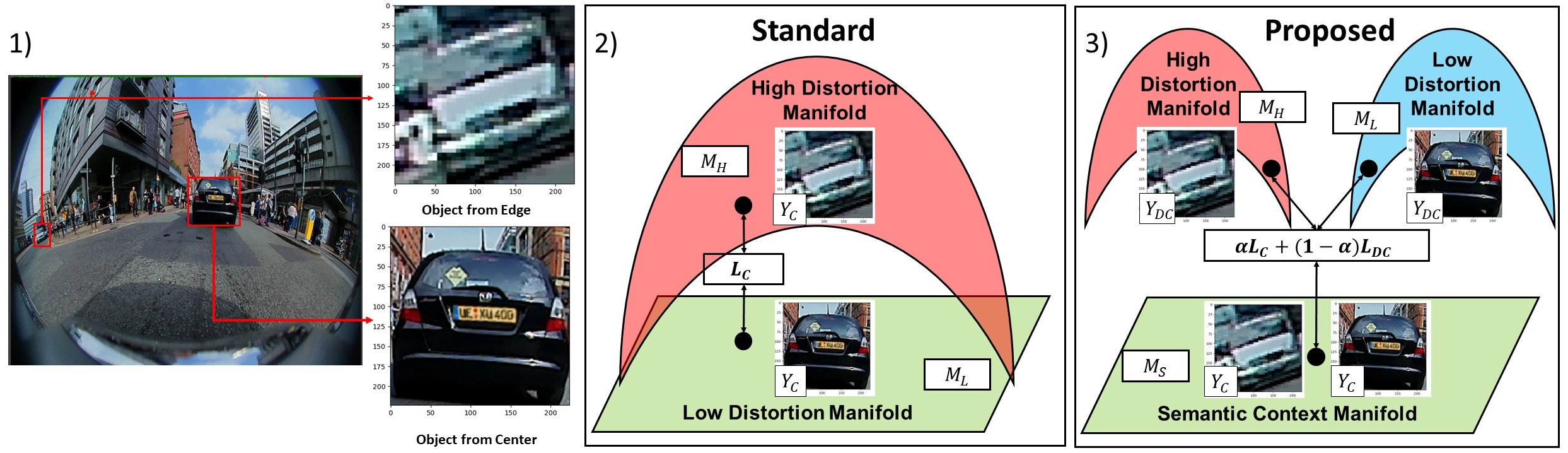}

\caption{1) This is an example fisheye object with associated car objects from the center and edge. 2) Previous methods view these objects as lying on separate manifolds that need to be corrected through a term that identifies them as belonging to the same semantic class $L_{C}$. 3) We propose an alternate view of the problem that views distortion and semantic context as belonging on separate sub-manifolds. The model is enabled to learn an intermediate representation that considers both concepts through a loss that enforces understanding of both the semantic class $L_{C}$ and a distortion class $L_{DC}$. }

\label{fig: manifolds}
\end{figure*}

\begin{figure}[t]
\centering
\includegraphics[scale=.35]{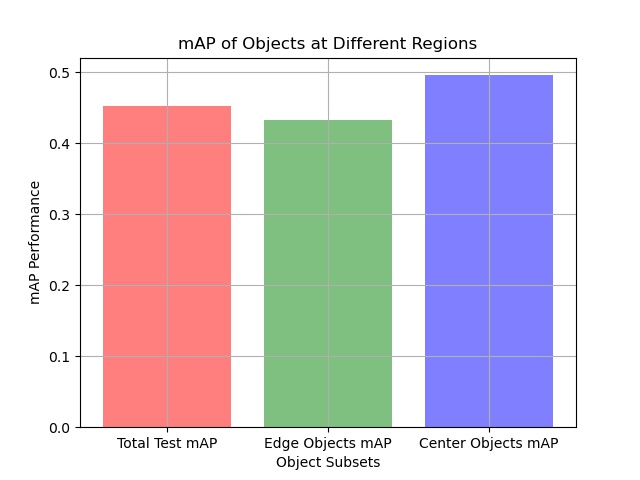}

\caption{This shows the mean average precision(mAP) for a toy experiment on the WoodScape \cite{yogamani2019woodscape} dataset. An object detector is trained using the YOLO v5 \cite{glenn_jocher_2022_7347926} framework. Then, we compute total mAP, mAP of just the objects at the edge of the image, and mAP of objects within the center of the image. }

\label{fig: edge_center_map}
\end{figure}

From a deep learning perspective, radial distortion introduces a plethora of issues because neural networks exhibit performance degradation outside of the pristine data setting \cite{temel2017cure}. Furthermore, most computer vision applications only consider data from narrow field of view cameras with mild radial distortion. As a result of these discrepancies, research has gone into developing deep learning approaches that maintain performance on fisheye data while simultaneously avoiding the sub-optimal process of rectifying the image. This research can be roughly divided into two sub-categories that we will refer to as model-centric and data-centric approaches. Model-centric refers to approaches \cite{rashed2020fisheyeyolo,ahmad2022fisheyehdk,miao2021improved} that attempt to change certain architectural features of a model with the intent of better conformation to an identified fisheye feature. Data-centric refers to approaches \cite{blott2018semantic,plaut20213d} that manipulate the available training data in an attempt to better generalize on the fisheye setting. While these approaches have seen varied levels of success, they are lacking in the sense that they are very specifically optimized for their task of interest. In other words, these approaches introduce improvements within their target application (monocular depth estimation, semantic segmentation, etc.), but they do not identify core properties that a model's representation of data should have in order to be tuned to the fisheye setting.

In this work, we address this research gap by introducing a representation-centric approach specifically designed for a general fisheye paradigm. Our perspective on this paradigm can be understood by the example presented in Figure \ref{fig: manifolds}. Within the first image, there are two cars: one that exists within the center of the frame and one that exists at the edge of the frame. It can be observed that the radial distortion property of fisheye data causes the car at the edge to exhibit a much higher level of distortion compared to the object taken from the center of the frame. To demonstrate the effect of this, we perform a toy experiment on the WoodScape \cite{yogamani2019woodscape} dataset to see how performance varies with respect to detecting center and edge located objects respectively. The results of this experiment on the Yolo v5 \cite{glenn_jocher_2022_7347926} architecture can be observed in Figure \ref{fig: edge_center_map} and it clearly shows a significant difference in performance of about .06 mAP between objects found in the center compared to those located at the edge of the image. One interpretation, shown in Figure \ref{fig: manifolds}, is that the distortion may cause the model to view these objects as coming from a high $M_{H}$ and low $M_{L}$ distortion manifold despite their membership within the same semantic class. While the data space reflects the distortion characteristic of the data, standard methods do not integrate distortion into the training paradigm and rely on just semantic labels $Y_{C}$ with an associated semantic-based loss $L_{C}$. However, previous work \cite{prabhushankar2022introspective,kwon2020backpropagated,kokilepersaud2022gradient} has shown that a model lacks generalization capability when the learnt representation does not reflect the underlying distribution of the data space. We argue in this work that the underlying distribution of fisheye data reflects not only semantic context or distortion alone, but a complex interaction between both. We visualize this perspective in part 3 of Figure \ref{fig: manifolds} where objects both exist on a semantic context manifold $M_{S}$ as well as separate distortion specific sub-manifolds $M_{H}$ and $M_{L}$. In this view of the problem, all objects have a label with respect to the semantic manifold $Y_{C}$ as well as labels that reflect their location within distortion space $Y_{DC}$. From this setup, it is then possible to train a model with a loss that integrates both the semantic characteristic $L_{C}$ and distortion characteristic $L_{DC}$ of the objects.  We implement such a framework by first extracting objects from fisheye data and using their distance from the center to assign distortion based class labels alongside their semantic class label. We then take advantage of contrastive learning approaches \cite{le2020contrastive} in order to explicitly enforce a model to learn a representation that reflects both the distortion and semantic characteristic of the object through a weighted contrastive loss $\alpha L_{C} + (1-\alpha)L_{DC}$. We then fine-tune the learnt representation within an object detection setting to empirically validate the approach. The target contributions of this work are: 

\begin{enumerate}
\vspace{-5mm}
  \item We introduce a representation-centric approach to training fisheye data based on using contrastive learning as a way to constrain the interaction between semantics and distortion.

  \item We perform an explicit analysis of this trade-off between being distortion-aware and semantically-aware within the context of an object detection setting.

  \item We compare against standard object detection and representation learning baselines to demonstrate the advantage of our approach. 
  
\end{enumerate}

\section{Related Works}
\subsection{Rectification Approaches}
One application of deep learning on fisheye data involves developing models to rectify the fisheye data with the goal of removing distortion. \cite{yin2018fisheyerecnet} introduced the use of a CNN as a means to extract features, parse the surrounding scene, and estimate the distortion parameters necessary to rectify the image. \cite{xue2019learning} built upon to this idea to derive distortion parameters by enforcing that straight lines maintain their straightness. Recent work \cite{yang2022fishformer} has explored the usage of transformer architectures to model domain shifts across distortions. \cite{chao2020self} demonstrated a self supervised approach for this task. While these works have shown good performance for the task of rectification, the downsides of a computational cost, rectification artifacts, and reduction in the field of view remain significant concerns. 

\subsection{Model-Centric}

Model-centric approaches refer to methods that introduce architectural changes in the hopes of performance improvements on their target task. Early work \cite{saito2011people} made use of probabilistic appearance models for pedestrian tracking. \cite{rashed2020fisheyeyolo} showed how changes to the output bounding box shape can improve the mIOU on distorted objects through better conformation to the distorted shape. Other work \cite{ahmad2022fisheyehdk} has introduced new types of feature extraction strategies such as the usage of a hyperbolic convolutional kernel. \cite{playout2021adaptable} demonstrated the usage of adaptable deformable kernels for semantic segmentation. \cite{chen2019smaller} introduced an additional feature pyramid block to detect smaller objects. Other ideas \cite{miao2021improved} showed how approximations to the domain of spherical data can work on fisheye data through the usage of an attention mechanism. While these architectural changes have shown performance improvements, it is unclear what aspects of fisheye data they are leveraging from the resultant features they extract. In our work, we make this explicit through our shaping of a representation space based on fisheye-centric principles of the interaction between distortion and semantics.  

\subsection{Data Centric}
Data-centric approaches  describe methods that intervene on the training data in the hopes of presenting the model with better views that are invariant to distortion. For example, \cite{blott2018semantic} introduced a set of data augmentations that were tuned to the fisheye setting. Additionally, \cite{plaut20213d} showed how training with geometric perspectives can enable better training views within the context of a 3D object detection task. The main issue with these approaches is that these augmentations are tuned to a specific task and it isn't clear what general augmentation principle is at work. Our work circumvents this by introducing how to create a general representation principles for fisheye data. 
\begin{comment}
\begin{figure}[t]
\centering
\includegraphics[scale=.55]{FIgures/class_num.png}

\caption{}

\label{fig: classes}
\end{figure}
\end{comment}

\subsection{Contrastive Learning}

The main idea behind contrastive learning approaches is to learn a lower-dimensional embedding space where similar pairs of images (positives) project closer to each other than dissimilar pairs of images (negatives). The manner in which these positives and negatives are defined  as well as their usage within the overall framework is what distinguishes contrastive learning methodologies from each other. Traditional approaches  like \cite{chen2020simple,chen2020improved,li2020prototypical} all choose positive instances by augmenting an image through some transformation and treating all other instances in the batch as the negative set. Within the domain of fisheye, \cite{ramchandra2022fisheyepixpro} has utilized existing contrastive learning approaches on fisheye data for the task of semantic segmentation. This work differs fundamentally from ours in the sense that we are proposing a contrastive learning approach specifically geared towards creating a fisheye specific representation space, rather than a generic space based on previous learning approaches. In order to do this, we leverage the supervised contrastive loss \cite{khosla2020supervised} where positives and negative instances are chosen on the basis of belonging to the same semantic category or not. In other words, an additional constraint is placed on the embedding space to guide further understanding of learning similar and dissimilar data points. This inspired recent work \cite{kokilepersaud2022gradient,kokilepersaud2022clinical,kokilepersaud2022volumetric} that generates labels using auxiliary information to shape a representation that is more appropriate for the application domain of medical and seismic data respectively. We introduce a way to make use of distortion auxiliary (distortion/semantic) information as a way to shape representations more suitable for the fisheye setting. 

\section{FishEye Image Analysis}
\subsection{Dataset}
The fisheye dataset utilized in this paper is the WoodScape \cite{yogamani2019woodscape} autonomous driving dataset. This dataset is collected using four surrounding fisheye cameras on a moving vehicle over a variety of urban scenes. It has over 8.2k images containing 5 different object categories: vehicle, pedestrian, bicycle, traffic light, and traffic sign. No public test set for this dataset is available, so the dataset was split into a training, validation, and test split of 80\%, 10\%, and 10\% respectively.  This specific dataset was chosen over the other fisheye datasets from \cite{kumar2022surround} for the following reasons. Firstly, WoodScape is the only real-world fisheye AV dataset with object bounding box labels. Other fisheye AV datasets are either unlabeled with respect to object bounding boxes or are entirely simulated. The non-AV fisheye datasets that are both real and have bounding boxes are insufficient to demonstrate the interaction between distortion and semantic classes. This is because the AV setting has a wide distribution of objects across the entire image, while non-AV datasets only have center-focused objects. This makes it hard to study the impact of objects closer to the edge where the distortion is highest.

\begin{figure}[t]
\centering
\includegraphics[scale=.4]{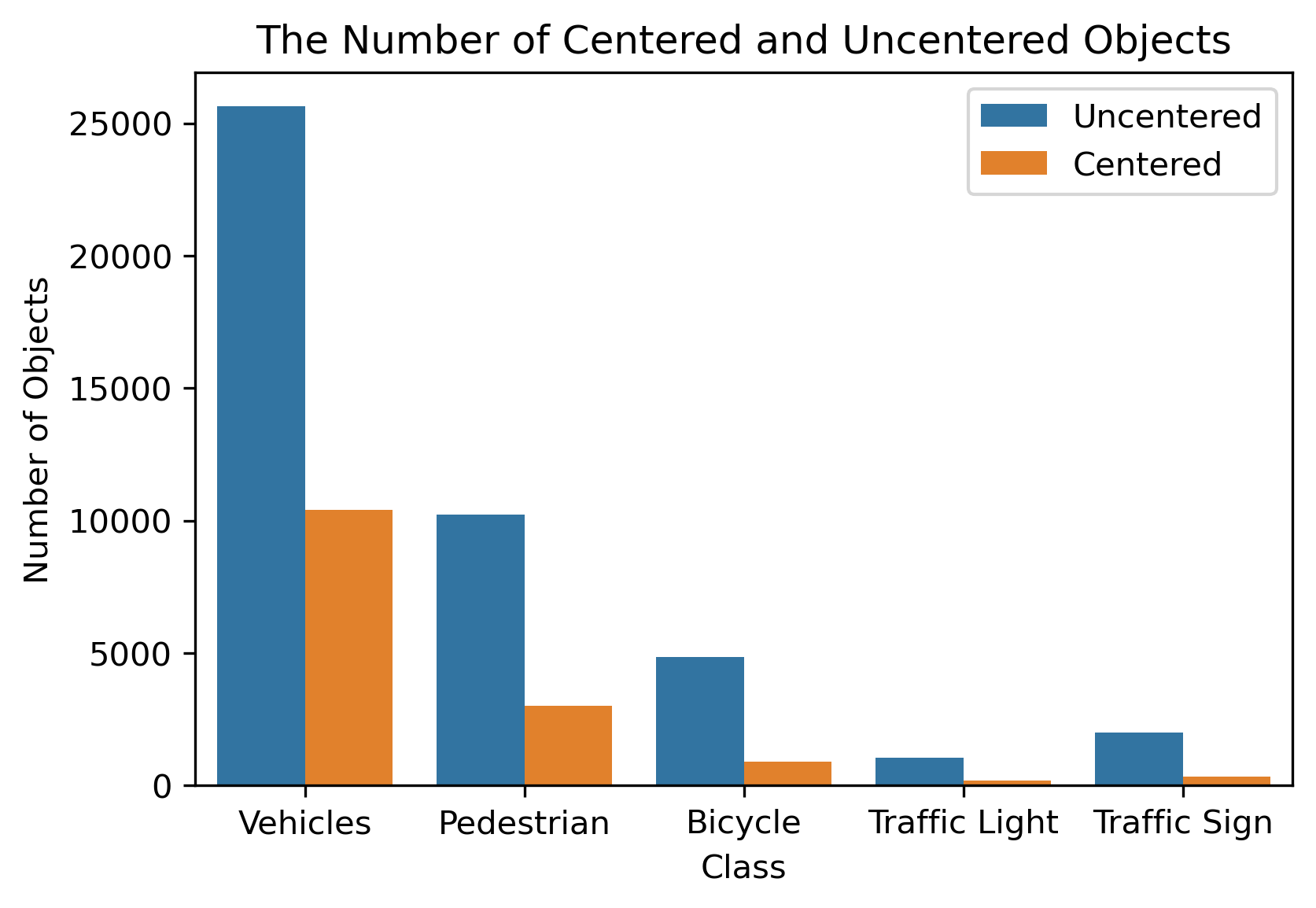}

\caption{This plot shows the number of objects located at the edge and the center for each of the classes in the WoodScape dataset.}

\label{fig: center_edge}
\end{figure}

\subsection{Statistics}
In Figure \ref{fig: edge_center_map}, we observe that the test set performance of objects located at the edge is lower than that of objects located in the center of the image. In this section, we build intuition regarding the interaction between object distortion and regional location. We provide analysis that demonstrate that the performance difference of Figure \ref{fig: edge_center_map} is due to the radial distortion of the fisheye images. To enable this analysis, we define two categories of objects - central and edge objects. All objects within an upper left image coordinate of (.25, .25) and lower right coordinate of (.75, .75) are central objects and objects outside this box are edge objects.

\begin{figure}[t]
\centering
\includegraphics[scale=.4]{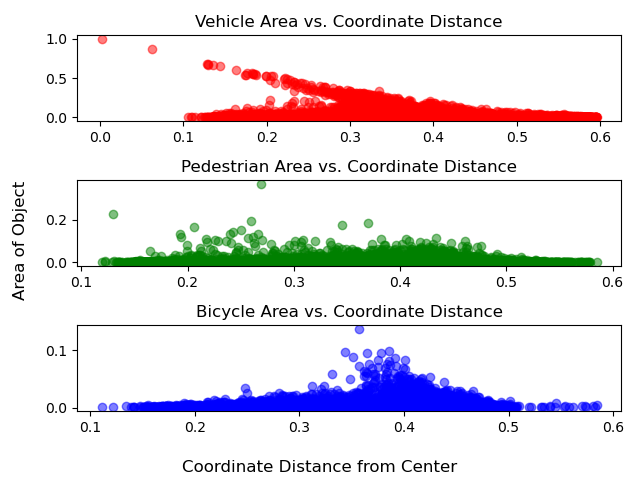}

\caption{This plot shows the relationship between the coordinate distance and object area for every object among the three most prevalent classes in the WoodScape dataset. Distance was computed as the mean squared error between the center coordinate of the image (.5,.5) and the object's center coordinate. Object area was computed by multiplying the height and width of the object's bounding box. The maximum possible distance is 0.707 and the lowest is 0.}

\label{fig: distance_area}
\end{figure}

\begin{figure}[t]
\centering
\includegraphics[scale=.4]{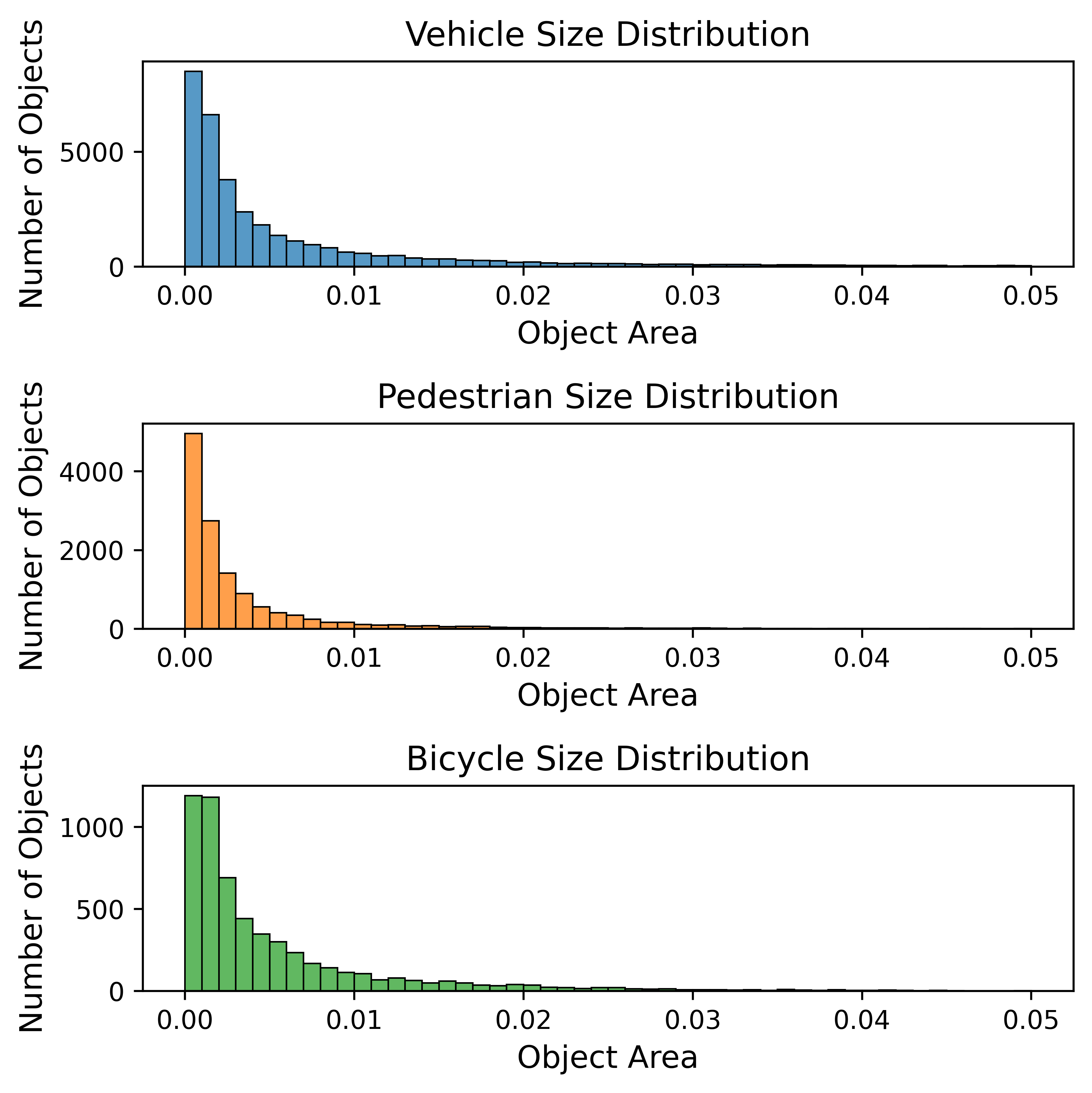}

\caption{This is a histogram of the number of objects at different sizes for each class. The area for each object is computed by multiplying the height and width for each objects bounding box coordinates. This is then ranked and binned to create this histogram.}

\label{fig: object_area}
\end{figure}

The first statistic we investigate is the distribution of the centered and edge objects across different classes in the training set. We can see from  Figure \ref{fig: center_edge} that the majority of objects are located closer to the edge of the image. This plot was generated by extracting the center coordinate for every object in every class and then using our definition of center and edge to delegate which location bin they belong to. This indicates that in Figure \ref{fig: edge_center_map} the model was not biased towards detecting central objects better because of a greater prevalence of objects to train on from the center of the image. Another statistic to investigate is the size of objects at different regions of the image. This is shown in Figure \ref{fig: distance_area} for the three most prevalent classes: vehicles, pedestrians, and bicycles. In this plot, every object's distance from the center is plotted against the area of their respective bounding box. We observe that the vast majority of objects in every class maintains a roughly similar size regardless of their distance from the center. This is further validated by the histogram in Figure \ref{fig: object_area} that shows the number of objects as a function of the object area for the majority classes. This histogram shows that most objects within each class have a similar size to each other. Together these plots show that the worse performance on the edges is not due to having smaller sized objects to detect compared to the center. Both regions have similarly sized objects with the difference being the higher radial distortion exhibited by the objects at the edge. 

\begin{figure*}[t]
\centering
\includegraphics[scale=.35]{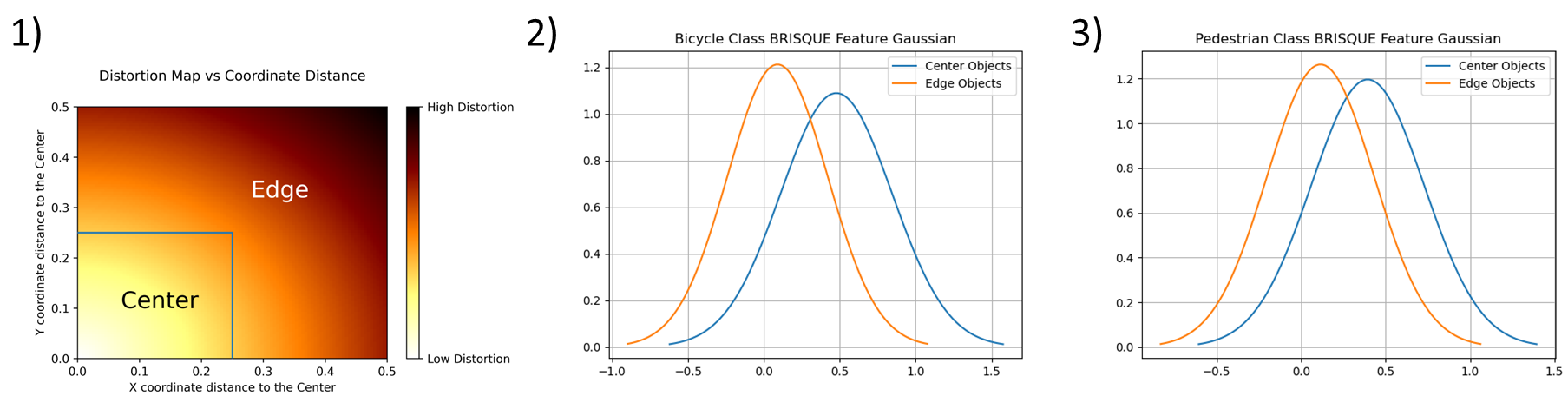}

\caption{This plot shows different statistics regarding distortion of center and edge objects. 1) This shows how distortion varies according to the fisheye polynomial distortion model $d(\rho) = a_{0} + a_{2}\rho^{2} + a_{3}\rho^{3} + a_{4}\rho^{4}$ where $\rho = \sqrt(x^{2} + y^{2}$ and $d(\rho)$ represents the associated distortion at $\rho$ distance. The parameters $a_{0} ... a_{4}$ are chosen from calibration files provided by the WoodScape dataset. We also show what x and y coordinate distances correspond to our definition of center and edge.  2) and 3) To produce these plots, every object for both the bicycle and pedestrian classes were passed through the BRISQUE\cite{mittal2012no} algorithm to produce a 36 x 1 feature vector that. The mean and standard deviation of this vector were takent to produce the associated Gaussian. }

\label{fig: iqa}
\end{figure*}

To further build on this idea, we attempt various ways in Figure \ref{fig: iqa} to quantify the distortion exhibited by objects on the edge in comparison to objects in the center. The first plot shows how distortion changes according to the radial distortion mathematical model described in \cite{miyamoto1964fish}. This shows the associated distortion should theoretically increase the further the away from the center of the image associated features are located. To evaluate this empirically, we also extract each object from both the pedestrian and bicycle classes and assign them as center and edge objects according to the conventions we introduce. We then compute BRISQUE \cite{mittal2012no} features that quantifies losses in "naturalness" due to distortions. This results in every image having an associated 36 x 1 feature vector. We compute a single value by averaging across this feature vector for each object. We compute the mean and standard deviation for this value across all objects in each class and associated regional location and plot the associated Gaussian. We observe that there is significant separation in BRISQUE features between the edge and center located objects. Specifically, this corresponds to a percent overlap of 18.55\% for the pedestrian class 10.90\% for the bicycle class which indicates that the distortion had some effect on the distributions of these objects.

\begin{comment}
To achieve this, we compute a BRISQUE \cite{mittal2012no} score for every object located in the center and at the edge and then average it for the most prevalent classes. This score is a no-reference image quality metric that attempts to quantify losses in "naturalness" due to the presence of distortions. Figure \ref{fig: iqa} shows that the BRISQUE score is higher on average for objects from the center compared to those from the edge. This shows that the objects display greater distortion the further they are located from the center. All of these statistics show that WoodScape object sizes, quantity, and location are not biased towards worse center performance and that the only major difference causing performance discrepancies is the interaction with the radial distortion from the fisheye projection.
\end{comment}

\begin{comment}
\begin{figure}[t]
\centering
\includegraphics[scale=.5]{FIgures/Objects_area.png}

\caption{}

\label{fig: object_area}
\end{figure}
\end{comment}

\vspace{-3mm}
\section{Methodology}

Our methodology follows three distinct steps: regional label extraction, followed by pre-training of a ResNet-18 network \cite{he2016deep} with a linear combination of contrastive losses, and finally fine-tuning the learnt representation with an object detection head.  The overall philosophy behind this approach is to enable the model to recognize both semantic and distortion related information within its representation space. The regional extraction provides us with the labels for distortion and the contrastive learning operates with a weighted loss that constrains the representations learnt in terms of both semantic and distortion related contexts.

\subsection{Regional Class Label Extraction}
 In order to train a model to recognize both semantic and distortion concepts, we need to acquire labels that reflect both. In the case of semantic information, the label files for every image identifies the class that each object belongs to. However, there isn't an explicit label that reflects the distorted nature of the object. In order to acquire this, we use the bounding box information of each object to receive the center coordinate of the object. This coordinate is important because the further the object from the center of the frame, the greater the distortion characteristics it exhibits. Therefore, it is necessary to define a threshold by which all objects outside of this threshold belong to a high-distortion class and all objects within this threshold belong to a lower distortion class. 
\begin{figure}[t]
\centering
\includegraphics[scale=.26]{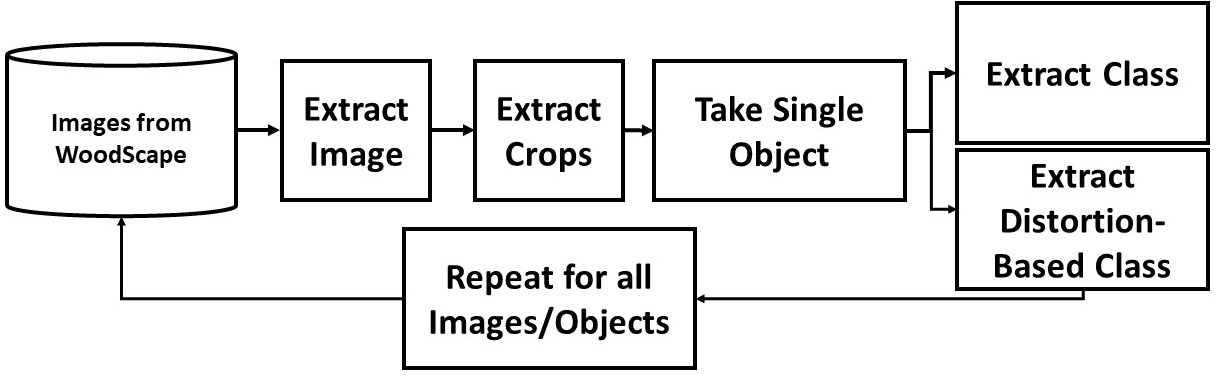}

\caption{This figure details the process by which objects are extracted and assigned an additional label based on distortion information. Objects from the WoodScape dataset are extracted and assigned their class label as well as a distortion class label based on whether the object belongs in the defined center region box or not. This is repeated for all objects to mine a training set of object patches with both semantic and distortion labels.}

\label{fig: region_extract}
\end{figure}
\begin{comment}
\begin{figure}[t]
\centering
\includegraphics[scale=.4]{FIgures/region_class_class.jpg}

\caption{After the extraction process of Figure \ref{fig: region_extract}, every object is labeled with both a class label and a distortion class label as shown in this figure.}

\label{fig: region_class_class}
\end{figure}
\end{comment}
With these considerations in mind, the process to define these distortion-based labels is detailed in Figure \ref{fig: region_extract}. For each image in the training set, we extract each individual object as its own image patch along with its associated class label and bounding box coordinates. Every object is immediately assigned its original class label, but it is also assigned an additional label to describe its distortion class. This is done by analyzing the center coordinate. If the center coordinate of the object belongs in the inscribed box with an upper left coordinate of (.25, .25) and lower right coordinate of (.75, .75) then we consider the object to belong close to the center of the image. In this case, we assign the object with the additional label of a lower distortion version of its class. If the center coordinates lies outside of this defined box, we assign the object with the label of a higher level distortion version of its class. This extraction and label assignment process is repeated across the training set to create a large pool of object patches with associated label information. For example, this means that every car object receives its semantic class of car as well as its appointment as a high or low distortion version of its class, such as highly distorted car and barely distorted car, resulting in 10 possible distortion classes due to two variants of each of the 5 classes. 
\subsection{Contrastive Pre-Training}
\begin{figure}[t]
\centering
\includegraphics[scale=.25]{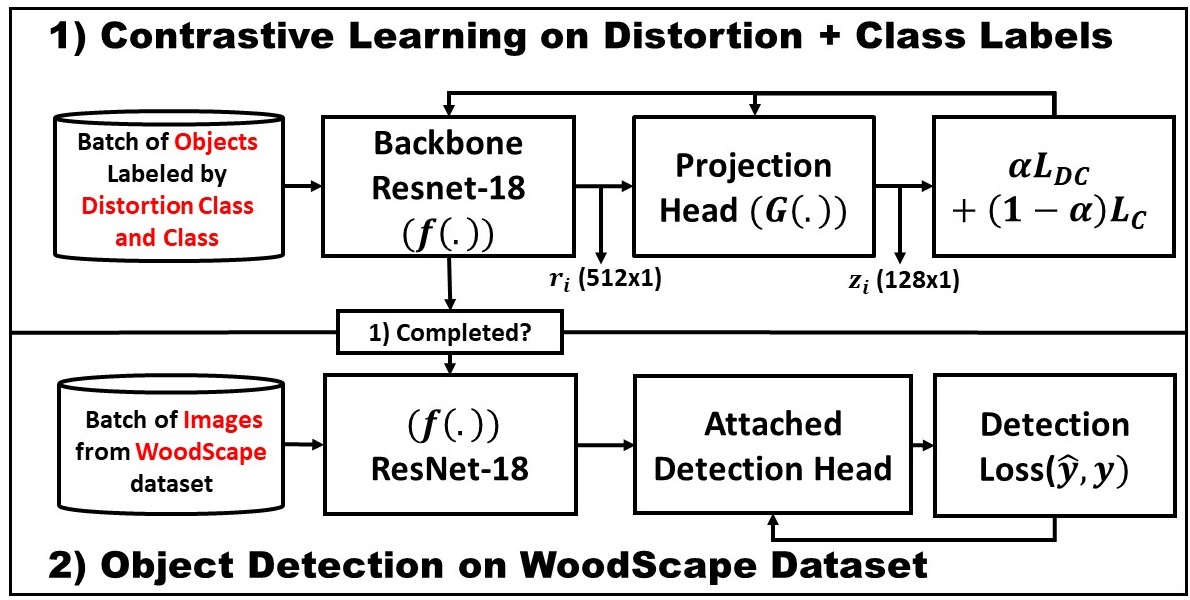}

\caption{This is the overall setup. 1) The backbone ResNet-18 model is trained using the combined contrastive loss on objects that incorporate both distortion and semantic class labels. 2) The model pre-trained with the contrastive loss in the previous step is transferred as the backbone for a YOLO v5 object detection setup. }

\label{fig: methodology_full}
\end{figure}

After disentangling the training set into object patches labeled with both a semantic and distortion class, we perform a contrastive learning objective that constrains the representation to consider both concepts. The overall block diagram of the proposed method is summarized in Figure \ref{fig: methodology_full}. Given an input batch of extracted objects $x_{k}$, associated distortion label $(y_{dk})$, and associated class label $(y_{ck})$ to form the triplet $(x_{k},y_{dk},y_{ck})_{k=1,...,N}$, we perform augmentations on the batch twice in order to get two copies of the original batch with $2N$ object patches and corresponding labels. These augmentations are random resize crop, random horizontal flips, random color jitter, and data normalization. 
This process produces a larger set $(x_{l},y_{dl},y_{cl})_{l=1,...,2N}$ that consists of two versions of each object patch that differ only due to the random nature of the applied augmentation. Thus, for every object patch $x_{k}$, distortion label $y_{dk}$, and class label $y_{ck}$  there exists two views of the image $x_{2k}$ and $x_{2k-1}$ and two copies of the labels that are equivalent to each other: $y_{2dk-1} = y_{2dk} = y_{dk}$ and $y_{2ck-1} = y_{2ck} = y_{ck}$.

From this point, we perform the first step in Figure \ref{fig: methodology_full}, where a linear combination of supervised contrastive losses is performed on the identified distortion and class based labels. The labeled augmented batch of object patches is forward-propagated through an encoder network $f(\cdot)$ that we set to be the ResNet-18 architecture \cite{he2016deep}. This results in a 512-dimensional vector $r_{i}$ that is sent through a projection network $G(\cdot)$, which further compresses the representation to a 128-dimensional embedding vector $z_{i}$. $G(\cdot)$ is chosen to be a multi-layer perceptron network with a single hidden layer. This projection network is utilized only to reduce the dimensionality of the embedding before computing the loss and is discarded after training. A supervised contrastive loss is performed on the output of the projection network in order to train the encoder network to have a weighted constraint based on both class and distortion labels. In this case, embeddings with the same class label are enforced to be projected closer to each other while embeddings with differing class labels are projected away from each other. At the same time, another loss enforces embeddings with the same distortion label to be projected closer to each other while embeddings with differing distortion labels are projected away from each other. This results in a class based supervised contrastive loss $L_{C}$ and a distortion class based supervised contrastive loss $L_{DC}$. The form of the distortion contrastive loss is shown as:  
$
    L_{DC} = \sum_{i\in{I}} \frac{-1}{|DC(i)|}\sum_{dc\in{DC(i)}}log\frac{exp(z_{i}\cdot z_{dc}/\tau)}{\sum_{a\in{A(i)}}exp(z_{i}\cdot z_{a}/\tau)}
$
where $i$ is the index for the object patch of interest $x_{i}$.
All positives $dc$ for object patch $x_{i}$ are obtained from the set $DC(i)$ and all positive and negative instances $a$ are obtained from the set $A(i)$. Set $DC(i)$ represents all other object patches in the batch with the same distortion class label $dc$ as the object patch of interest $x_{i}$ while set $A(i)$ refers to every other element in the same batch as $x_{i}$. Additionally, $z_{i}$ is the l2-normalized embedding for the object patch of interest. $z_{dc}$ represents the embedding for the distortion class positives, and $z_{a}$ represents the embeddings for all positive and negative instances in the set $A(i)$. $\tau$ is a temperature scaling parameter that is set to .07 for all experiments.  The loss function operates in the embedding space where the goal is to maximize the cosine similarity between embedding $z_{i}$ and its set of distortion class positives $z_{dc}$. The class based contrastive loss follows the same setup except positives are chosen on the basis of only class information. In order to weight the influence that each term has on shaping the representation learnt by the model we introduce an $\alpha$ parameter. This weighting between each contrastive loss can be represented by: $L_{total} = \alpha L_{DC} + (1-\alpha)L_{C}$. 
In this way, we are creating a linear combination of losses from different label distributions for the same object patch.

\begin{figure*}[t]
\centering
\includegraphics[scale=.3]{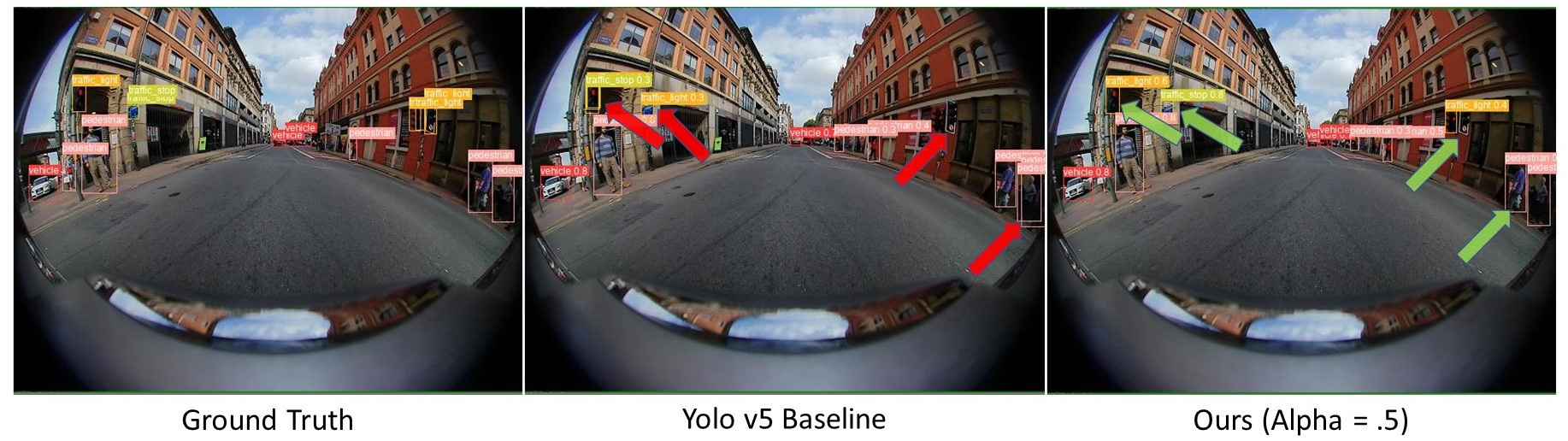}

\caption{This is a visual comparison between objects detected by the Yolo v5 baseline architecture and our method with an alpha parameter of .5. We include red and green arrows to highlight where our method did better than the baseline Yolov5.}

\label{fig: visual}
\end{figure*}

\subsection{Object Detection Fine-Tuning}

After training the encoder on objects with  a combined contrastive loss, we move to the second step in Figure \ref{fig: methodology_full} where the weights of the encoder are transferred into the backbone of the object detection setup and a YOLO v5 object detection head \cite{glenn_jocher_2022_7347926} is appended to the output of the encoder. The images from the WoodScape training set are input into this setup and the model is trained to perform standard object detection. In this way, we leverage knowledge learnt from constraining representations based on class and distortion information in order to improve performance for the task of object detection. 

\section{Results}
\subsection{Training Details}
The hyperparameters utilized can be divided into those applied for the contrastive learning step and that meant for the object detection fine-tuning. During contrastive learning, we set a batch size of 64 and training was performed for 25 epochs. A stochastic gradient descent optimizer was used for contrastive pre-training with a learning rate of .001, weight decay of .0001, and momentum of .9. The applied augmentations are random resize crop, random horizontal flips, random color jitter, and data normalization to the mean and standard deviation of the Woodscape dataset. The comparison methods of SimCLR\cite{chen2020simple}, Moco v2\cite{chen2020improved}, and PCL \cite{li2020prototypical} were trained in the same manner with certain hyper-parameters specific to each method. Specifically, Moco v2 was set to its default queue size of 65536. Additionally, PCL has hyper-parameters specific to its clustering step, but the original documentation made these parameters specific to the Imagenet \cite{deng2009imagenet} dataset on which it was originally built for. To fit these parameters to our setting, the clustering step was reduced in size.

%%%%%%%%%%%%%%%%%%%%%%%%% Discrete Variation

Object detection fine-tuning on top of the contrastively trained representation space follows many of the same hyperparameter choices as in the original Yolo v5 training setup. The main points to note are resizing all images to a size of 640 x 640, a training time of 100 epochs, a chosen batch size of 32, and a stochastic gradient descent optimizer with a learning rate of .01 that follows a cosine learning rate scheduler as training progresses. Further details of the architecture of the object detection head and its parameters can be found in the original YOLO v5 codebase \cite{glenn_jocher_2022_7347926}. 

\subsection{Alpha Parameter Analysis}
\begin{figure}[t]
\centering
\includegraphics[scale=.4]{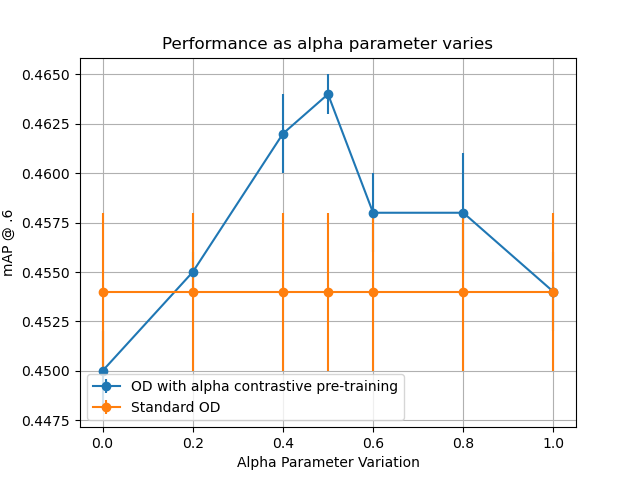}

\caption{The plot shows the effect of varying the alpha parameter on the combined contrastive loss. We compare this against a standard object detection (Standard OD) setup that does not benefit from access to contrastive pre-training of the backbone network.}

\label{fig: alpha_var}
\end{figure}

In order to get a sense of the trade-off between optimizing for distortion and semantic information, we vary the alpha parameter on the combined contrastive loss $L_{total} = \alpha L_{DC} + (1-\alpha)L_{C}$  that we introduce. In this way, we can observe how shaping the representation with respect to each loss term effects downstream performance once fine-tuning for the object detection task is performed. We observe in Figure \ref{fig: alpha_var} that the choice of alpha has a significant impact on the downstream performance of the model. Specifically, we note that when $\alpha = .5$, the performance is highest and when $\alpha = 0$ or $\alpha = 1$ the performance is no better than the standard object detection baseline. In other words, when the learnt representation is forced to consider both the distortion and semantic information equally, the performance is much better than when the model is subjected to each alone. This trend holds as performance decreases on either side of $\alpha=.5$ as the $\alpha$ value increases or decreases. This result is significant because it validates the idea that a good representation space for fisheye data is one that reflects this interaction between distortion and semantic information. In particular, an $\alpha=.5$ yielding the best performance suggests that both are equally important for shaping a good representation space. A possible reason for this performance increase relates to the work of \cite{fu2022details} where the authors show that a representation space should not map all instances of a class to the same point, but rather uniformly distribute them across the surface of a hypersphere based on implicit sub-classes within each higher order class. Within our setting this means that not only should objects of the same class map close to each other, but also be distributed based on distortion characteristics. We also observe visually in Figure \ref{fig: visual} that our method is able to more accurately detect objects compared to the standard baseline method. Additionally, we see a performance boost for both center and edge mAP performance in Table \ref{tab:edge_center}.

\begin{figure}[t]
\centering
\includegraphics[scale=.4]{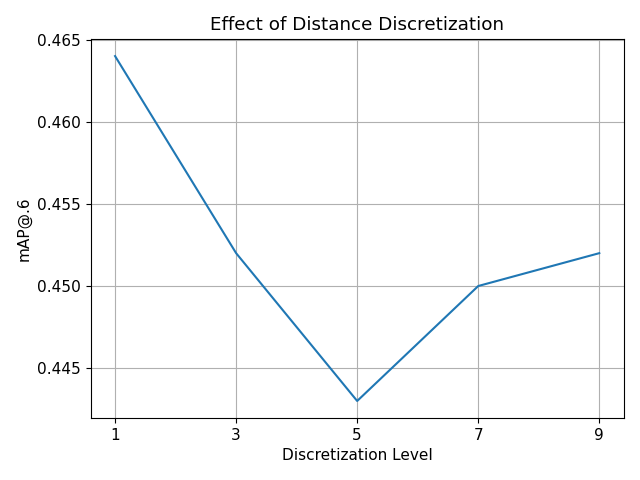}

\caption{In this plot, the definition of objects that exhibit high and low distortion are changed to include multiple possible regions of distortion. Discretization level indicates the number of distortion regions generated. The performance shown is how well each contrastive pre-training did when the distortion class is re-defined with the larger number of distortion levels.}

\label{fig: discrete_analysis}
\end{figure}

\begin{table}[]
\centering
\begin{tabular}{@{}cccc@{}}
\toprule
\multicolumn{4}{c}{Edge and Center Study}   
\\ \midrule
\multicolumn{1}{c}{Method} & \multicolumn{1}{c}{Total mAP} & \multicolumn{1}{c}{Edge mAP@.6} & \multicolumn{1}{c}{Center mAP@.6} \\
\midrule
Baseline Yolo v5  & .453 & .433  & .496  \\  

Ours (alpha = .5)  & \textbf{.464} & \textbf{.445}  & \textbf{.505} \\

 \bottomrule
\end{tabular}
\vspace*{.9mm}
\caption{We observe a higher performance on the edge and center objects compared to the baseline approach.}
\label{tab:edge_center}
\end{table}
We also compare our approach at $\alpha=.5$ to other representation approaches in Table \ref{tab:representation}. Our approach beats all other contrastive learning strategies that only consider a simple augmentation as a means of constraining the representation space. We also note that this improvement in performance is consistent across different bounding box intersection over union (IOU) thresholds. This indicates that  the resultant model produces higher quality bounding boxes compared to the other methods. Part of the reason for this improvement is that traditional approaches choose positives and negatives that are less reflective of the data distribution. For example, by choosing the only positive pair as an augmentation and every other point in the batch as a negative, this leads to situations where the negatives consist of points that should be positives. This is avoided in our approach through having to a much more diverse pool of positives based on both class and distortion related considerations.

%%%%%%%%%%%%%%%%%%%%%%%%% Representation Comparison
\begin{table}[]
\centering
\begin{tabular}{@{}cccc@{}}
\toprule
\multicolumn{4}{c}{Comparison with other Representation Learning Approaches}   
\\ \midrule
\multicolumn{1}{c}{Method} &\multicolumn{1}{c}{mAP@.5} & \multicolumn{1}{c}{mAP@.6} & \multicolumn{1}{c}{mAP@.7} \\
\midrule
Standarad OD  & $.457$ & $.454$ & $.433$  \\  
\midrule
PCL  \cite{li2020prototypical}     & $.458$ & $.457$ & $.441$         \\

SimCLR \cite{chen2020simple}      & $.453$ & $.451$ & $.430$              \\
Moco v2 \cite{chen2020improved}    & $.462$ & $.459$ & $.444$              \\ \bottomrule
Ours ($\alpha$=.5) & $\textbf{.467}$ & $\textbf{.464}$ & $\textbf{.448}$              \\

 \bottomrule
\end{tabular}
\vspace*{.9mm}
\caption{This shows the performance of our best contrastive pre-training model against other well known representation learning approaches. Performance is reported at different IOU thresholds for the mAP metric.}
\label{tab:representation}
\end{table}

\begin{table}[]
\centering
\begin{tabular}{@{}cc@{}}
\toprule
\multicolumn{2}{c}{Comparison with Different Edge/Center Box Definitions}   
\\ \midrule
\multicolumn{1}{c}{Box Definition} & \multicolumn{1}{c}{mAP@.6} \\
\midrule
Standard Box  & .464  \\  

Large Box  & .431 \\ 

Small Box  & .424 \\

 \bottomrule
\end{tabular}
\vspace*{.9mm}
\caption{We perform an ablation study of varying the definition of the boundary between center and edge objects. Standard box refers to the size we use in the paper of an upper left coordinate of (.25,.25) and lower right coordinate of (.75,.75). Large Box refers to  an edge/center boundary box defined by the coordinate pair (.1,.1) and (.9,.9). Small box efers to  an edge/center boundary box defined by the coordinate pair (.33,.33) and (.66,.66) }
\label{tab:box_size}
\end{table}

%%%%%%%%%%%%%%%%%%%%%%%%%%% Contrastive Learning Ablation Study
\begin{table}[]
\centering
\begin{tabular}{@{}cccc@{}}
\toprule
\multicolumn{4}{c}{Contrastive Learning Hyperparameter Study}   
\\ \midrule
\multicolumn{1}{c}{Batch Size} & \multicolumn{1}{c}{mAP@.6} & \multicolumn{1}{|c}{Temperature} & \multicolumn{1}{c}{mAP@.6} \\
\midrule
32  & .457 & .07  & .464  \\  

64  & .464 & 0.2  & .458 \\ 

128  & .458 & 0.5  & .449 \\

 \bottomrule
\end{tabular}
\vspace*{.9mm}
\caption{We study the effect of variation of batch size with a fixed temperature of .07 as well as variation of temperature with a fixed batch size of 64. Our chosen parameters of a batch size of 64 and temperature of .07 lead to the best performance. }
\label{tab:contrastive_hyper}
\end{table}

%%%%%%%%%%%%%%%%%%%%%%%%%%%%%%%%%%%%%%%% Architecture Ablation Study
\begin{table}[]
\centering
\begin{tabular}{@{}cccc@{}}
\toprule
\multicolumn{4}{c}{Architecture Study}   
\\ \midrule
\multicolumn{1}{c}{Method} & \multicolumn{1}{c}{Yolo v5 \cite{glenn_jocher_2022_7347926}} & \multicolumn{1}{c}{RetinaNet \cite{lin2017focal}} & \multicolumn{1}{c}{EfficientDet \cite{tan2020efficientdet}} \\
\midrule
Baseline  & .453 & .229  & .241  \\  

Ours (alpha = .5)  & \textbf{.464} & \textbf{.250}  & \textbf{.251} \\

 \bottomrule
\end{tabular}
\vspace*{.9mm}
\caption{We study the effect of training on different families of architectures under the baseline method and ours with the best empirical alpha parameter. }
\label{tab:architecture}
\end{table}

%%%%%%%%%%%%%%%%%%%%%%%%%%%%%%%%%%%%%%%%%%%%%%

\begin{comment}

%%%%%%%%%%%%%%%%%%%%%%%%% Model Variation
\begin{table}[]
\centering
\begin{tabular}{@{}cccs@{}}
\toprule
\multicolumn{4}{c}{Comparison with Other Models}   
\\ \midrule
\multicolumn{1}{|c|}{Method} & \multicolumn{1}{c|}{R-18} & \multicolumn{1}{c|}{R-50} \\\midrule
Standard Object Detection             \\
\midrule
PCL  \cite{li2020prototypical}          \\

SimCLR \cite{chen2020simple}              \\
Moco v2 \cite{chen2020improved}            \\ \bottomrule
Ours ($\alpha$=.5)         \\

 \bottomrule
\end{tabular}
\vspace*{.9mm}
\caption{}
\label{tab:models}
\end{table}
\end{comment}
\subsection{Experiment Variation Studies}

We also study different ways in which to define the distorted and clean image regions, different architectures, and different chosen contrastive learning hyperparameters. Within our work so far, we have defined a box by which all objects inside this box are considered low distortion and all objects outside this box are considered high distortion labels. In Table \ref{tab:box_size} we explore the effect of a larger and smaller 
definition of this boundary. We observe in both cases performance degrades compared to our standard box based on the midpoint distance between the center and upper left corner of the image. Another way of defining these regions is to discretize the distance from the center as a series of ranges by which each range of values would denote a separate distortion class. To study this possibility, we discretize the distance from the center into $l$ different distortion levels and train the contrastive learning setup in the same way discussed previously with the difference being the addition of additional distortion classes for every semantic class. We then fine-tune the representation trained in this manner for the task of object detection on the WoodScape dataset and report the results in Figure \ref{fig: discrete_analysis}. We observe that any level of discretization beyond the low and high levels we introduced leads to a substantial drop-off in performance in terms of mAP. Part of the reason for this is that distortion in fisheye data doesn't change at a fast enough rate that would cause substantial differences between distortion levels. Therefore, from a contrastive learning point of view, the loss does not have significant enough differences in terms of features to contrast between. 

We also ensure that our methods works in other object detection frameworks. We show a performance improvement over baseline RetinaNet \cite{lin2017focal} and EfficientDet \cite{tan2020efficientdet} in Table \ref{tab:architecture} when integrating our approach into the backbone pre-training of each network with an alpha weighting of $.5$. Additionally, we analyze the sensitivity of our approach with respect to batch size and temperature scaling in Table \ref{tab:contrastive_hyper}. As expected a higher temperature leads to poorer performance, as observed in \cite{khosla2020supervised}, and an optimal batch size choice results in the best performing setting. 
\section{Conclusion}
In this work, we investigate how a contrastive learning methodology can be used to enforce a model's representation space to reflect the distortion and semantic interaction inherent within fisheye data. We show how our method reflects this interaction through experiments that vary the alpha parameter during contrastive pre-training. Additionally, further experiments that compare against different representation learning strategies and discretization levels shows the introduced strategy out-performs existing approaches as well as standard object detection. We conclude from these experiments that a quality representation space is one the reflects the features of the data on which it is trained on. In this case, our models are better able to overcome the fisheye radial distortion by being allowed to integrate this information within the training process.

\bibliographystyle{IEEE}
\bibliography{ref}

\begin{IEEEbiography}[{\includegraphics[width=1in,height=1in,clip,keepaspectratio]{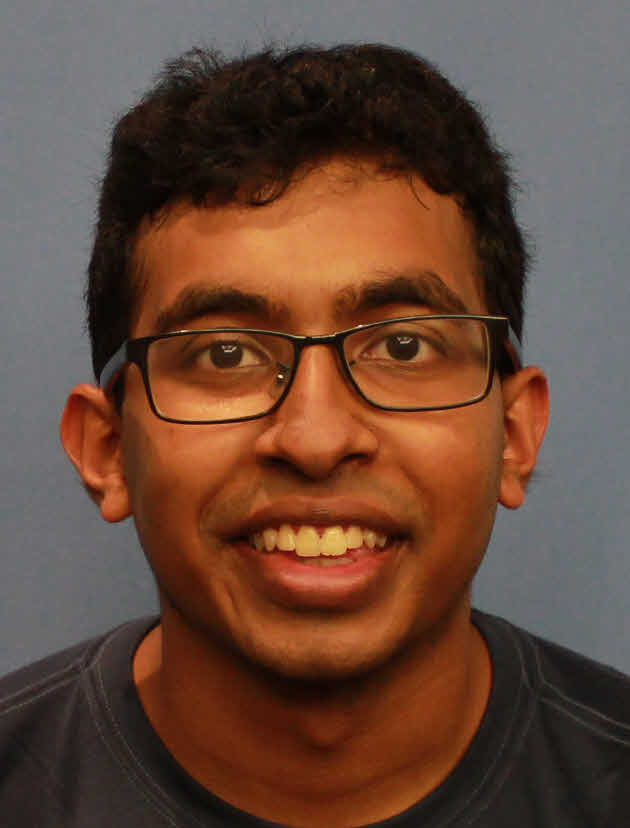}}]{Kiran Kokilepersaud}{\space} is a Ph.D. student in electrical and computer engineering at the Georgia Institute of Technology (Georgia Tech), Atlanta, Georgia, 30332, USA. He is currently a Graduate Research Assistant in the School of Electrical and Computer Engineering at the Georgia Institute of Technology in the Omni Lab for Intelligent Visual Engineering and Science (OLIVES) lab. He is a recipient of the Georgia Tech President's Fellowship for excellence amongst incoming Ph.D. students. His research interests include digital signal and image processing, machine learning, and its associated applications within the medical field.
\end{IEEEbiography}

\begin{IEEEbiography}[{\includegraphics[width=1in,height=1in,clip,keepaspectratio]{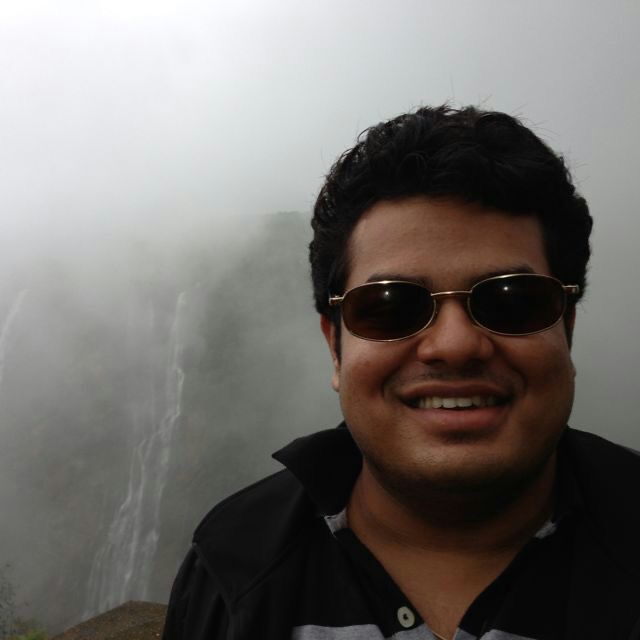}}]{Mohit Prabhushankar}{\space} received his Ph.D. degree in electrical engineering from the Georgia Institute of Technology (Georgia Tech), Atlanta, Georgia,
30332, USA, in 2021. He is currently a Postdoctoral Researcher and Teaching Fellow in the School of Electrical and Computer Engineering at the Georgia Institute of Technology in the Omni Lab for Intelligent Visual Engineering and Science (OLIVES) lab. He is working in the fields of image processing, machine learning, explainable and robust AI, active learning, and healthcare. He is the recipient of the Best Paper award at ICIP 2019 and Top Viewed Special Session Paper Award at ICIP 2020. He is the winner of the Roger P Webb ECE Graduate Research Excellence award in 2022. He is an IEEE Member.
\end{IEEEbiography}

\begin{IEEEbiography}[{\includegraphics[width=1in,height=1in,clip,keepaspectratio]{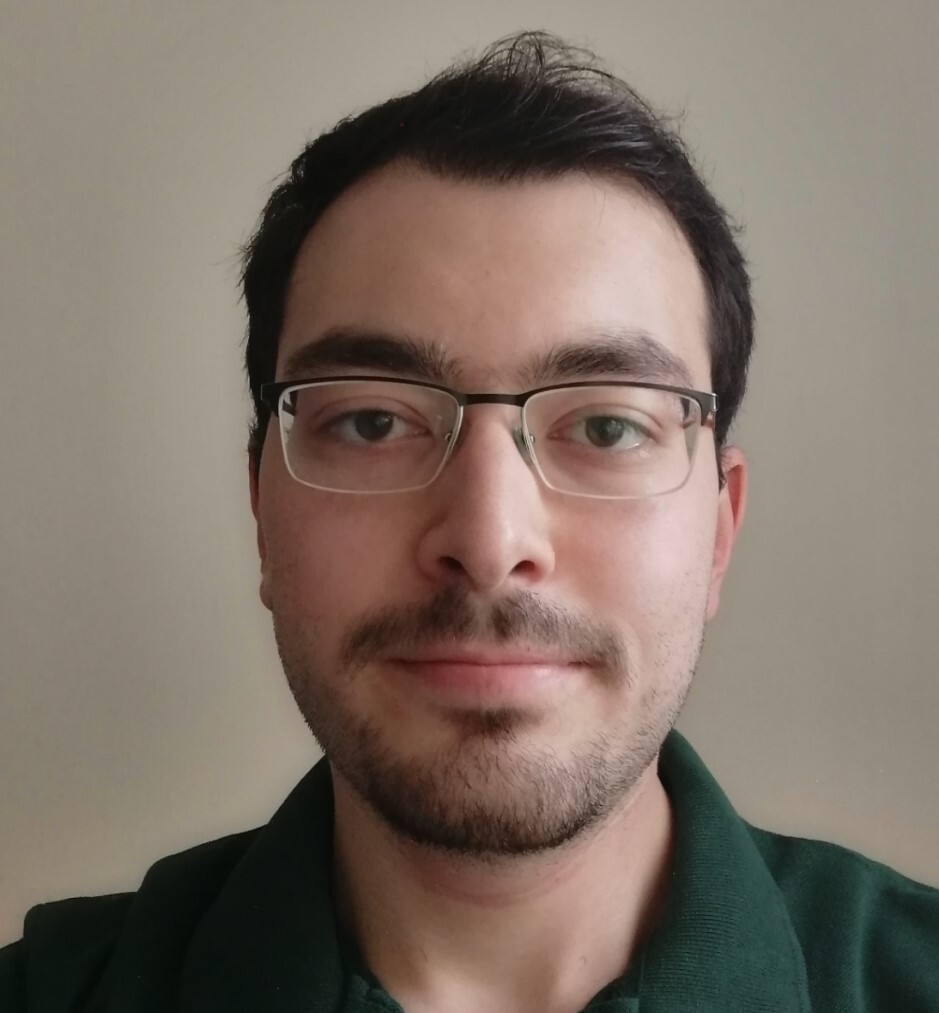}}]{Yavuz Yarici}{\space} is a Ph.D. student in the Omni Lab for Intelligent Visual Engineering and Science (OLIVES) at the Georgia Institute of Technology. He is a recipient of a Georgia Tech ECE Fellowship. His research interests include machine learning, computer vision, and image processing. Prior to Georgia Tech, he received his B.Sc degree in Electrical and Electronics Engineering from Bilkent University in Turkey.
\end{IEEEbiography}

\begin{IEEEbiography}[{\includegraphics[width=1in,height=1in,clip,keepaspectratio]{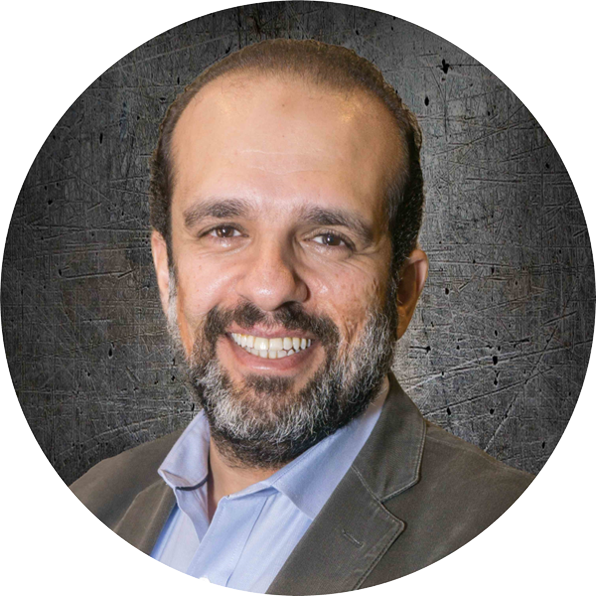}}]{Ghassan AlRegib}{\space} is the John and Marilu McCarty Chair Professor in the School of Electrical and Computer Engineering at the Georgia Institute of Technology. He was the recipient of the ECE Outstanding Junior Faculty Member Award, in 2008, and the 2017 Denning Faculty Award for Global Engagement. His research group, the Omni Lab for Intelligent Visual Engineering and Science (OLIVES) works on research projects related to explainable machine learning, robustness in intelligent systems, interpretation of subsurface volumes, and expanding healthcare access and quality. He has participated in several service activities within the IEEE and served on the editorial boards of several journal publications. He served as the TP co-Chair for ICIP 2020 and GlobalSIP 2014. He served as expert witness on several patents infringement cases and advised several corporations on both technical and educational matters. He is an IEEE Fellow.  
\end{IEEEbiography}

\begin{IEEEbiography}[{\includegraphics[width=1in,height=1in,clip,keepaspectratio]{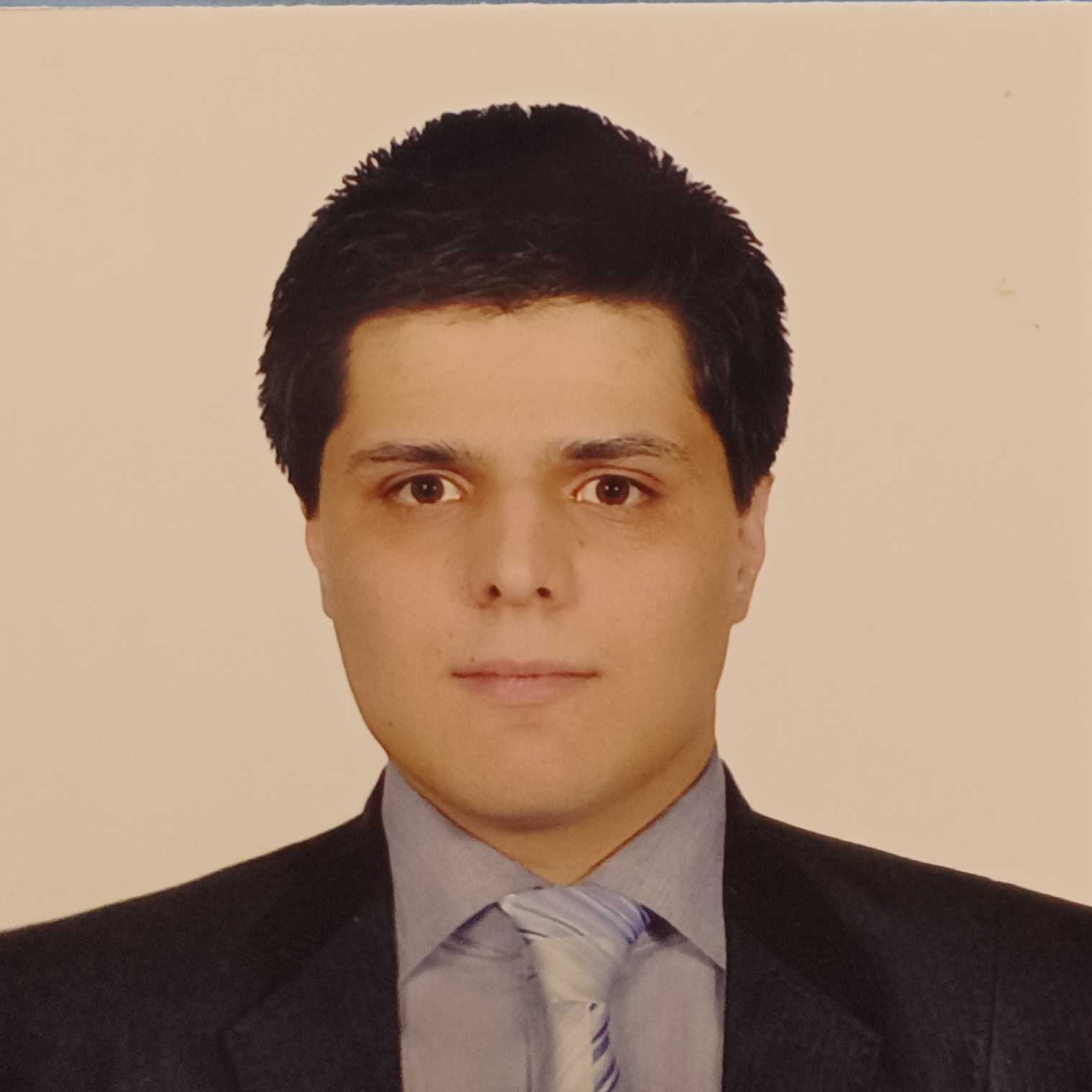}}]{Armin Parchami}{\space} ecieved his B.E. in Software Engineering and M.Sc. in Artificial Intelligence from Bu-Ali Sina University and then he received his Ph.D. in Computer Science from UTA in 2017. His dissertation was on single shot face recognition using deep learning algorithms for security applications. Between 2017 and 2022, he was working at Ford on level 4 autonomous vehicles. He is currently managing the perception team at Ford ADAS developing perception algorithms for L2+ autonomy. His current research interests include monocular 3D object detection, active learning, and wide baseline sensor fusion.
\end{IEEEbiography}

\end{document}